\documentclass{article}
\pdfoutput=1

\PassOptionsToPackage{numbers, compress}{natbib}

\usepackage[final]{neurips_2022}

\usepackage{float}          %

\usepackage[utf8]{inputenc} %
\usepackage[T1]{fontenc}    %
\usepackage{hyperref}       %
\usepackage{url}            %
\usepackage{booktabs}       %
\usepackage{amsfonts}       %
\usepackage{amsmath}        
\usepackage{nicefrac}       %
\usepackage{microtype}      %
\usepackage{xcolor}         %
\usepackage{graphicx}
\usepackage{caption}
\usepackage{subcaption}
\usepackage{bbm}
\usepackage{multirow}
\usepackage{cleveref}
\usepackage{amssymb}
\usepackage{xurl}
\usepackage[title]{appendix}
\DeclareMathOperator*{\argmax}{arg\,max}

\DeclareMathOperator*{\kargmax}{\mathit{k}-arg\,max}

\newcommand\x{\mathbf{x}}
\newcommand\y{\mathbf{y}}
\renewcommand\c{\mathbf{c}}
\newcommand\C{\mathbf{C}}
\newcommand\Henc{\mathbf{H}_{\texttt{Enc}}}

\newcommand*\samethanks[1][\value{footnote}]{\footnotemark[#1]}

\title{Decoupled Context Processing for \\ Context Augmented Language Modeling}

\author{%
 Zonglin Li \thanks{Equal contribution}~ \\
 Google Research, New York \\
 \texttt{lizonglin@google.com} \\
 \And
 Ruiqi Guo \samethanks~ \\
 Google Research, New York \\
 \texttt{guorq@google.com} \\
 \And
 Sanjiv Kumar \\
 Google Research, New York \\
 \texttt{sanjivk@google.com} \\
}

\begin{document}

\maketitle

\begin{abstract}
Language models can be augmented with a context retriever to incorporate knowledge from large external databases. By leveraging retrieved context, the neural network does not have to memorize the massive amount of world knowledge within its internal parameters, leading to better parameter efficiency, interpretability and modularity. In this paper we examined a simple yet effective architecture for incorporating external context into language models based on decoupled \texttt{Encoder-Decoder} architecture. We showed that such a simple architecture achieves competitive results on auto-regressive language modeling and open domain question answering tasks. We also analyzed the behavior of the proposed model which performs grounded context transfer. Finally we discussed the computational implications of such retrieval augmented models.
\end{abstract}

\section{Introduction}
Transformers have proven to be powerful language models that capture an impressive amount of world knowledge in its internal parameters and generalize to a variety of downstream tasks~\cite{vaswani2017attention,raffel2019t5}.
Recently, there has been a lot of success in improving language model quality by increasing the number of parameters in transformers, often on the order of hundreds of billion~\cite{chowdhery2022palm,rae2021gopher,smith2022mtnlg,brown2020gpt3}.
However, the scaling of model size also contributes to the exponential rise of the computation costs, both in terms of the number of accelerators needed and energy consumption~\cite{patterson2021carbon}.

To overcome the exponential increase in the number of parameters, one natural idea is to utilize information retrieved from an external source such as a massive external database, therefore freeing the neural network from having to memorize world knowledge.
To this end, researchers proposed multiple context augmented language model architectures~\cite{khandelwal2019knnlm,guu2020realm,borgeaud2021retro,yogatama2021spalm,lewis2020rag}. Such architecture typically has two components: a retriever that embeds the input sequence and retrieves relevant context from external source through vector similarity search; a neural network that integrates both the input and retrieved external context into the prediction of target sequence, formally:

\begin{equation}
P(\y| \x, \C=Retrieve(\x, \mathcal{D}); \mathbf{\theta}) \ge P(\y| \x, \mathbf{\theta'}) \label{eq:1}
\end{equation}

Here, $\C=\{\c\}$ is a set of context retrieved from the external database $\mathcal{D}$. $\mathbf{\theta'}$ is a self-contained language model which predicts target sequence $\y$ based solely on the input $x$ whereas $\mathbf{\theta}$ corresponds to the context augmented language model which incorporates both the input $\x$ and the retrieved context $\C$.

One of the challenges for such context augmented language model is the computational cost of context retrieval and incorporation, especially when multiple pieces of context is present or the context sequence is long.
In this paper, we propose a computationally efficient architecture for incorporating context based on vanilla \texttt{Encoder-Decoder}, which decouples the encoding of context and the prediction of target sequence.
We show that the model with such a simple architecture is competitive when compared with customized mechanisms such as Chunked-Cross-Attention~\cite{borgeaud2021retro} on language modeling score (as measured by bits-per-byte, BPB), while being more efficient in terms of parameter count and computation cost. 
Then, we define metrics to measure the utility of the retrieved context and use it to guide the training of the retriever.
We further show competitive results on downstream tasks of question answering, and demonstrate that the model takes advantage of the retrieved context without memorizing facts within its internal parameters.
Finally, we study the implication of context retrieval in terms of retrieval latency, accuracy and computation cost.

To summarize the main contributions of this article:
\begin{itemize}
    \item Proposed a novel Encoder-Decoder based architecture for incorporating retrieved external context, which decouples context encoding from language model inference.
	\item Demonstrated the competitive results of the proposed model on both the auto-regressive language modeling task and the open domain question answering task. 
    \item Analyzed model behavior by understanding how context improves language modeling prediction on tokens with different linguistic properties and how the model performs grounded context transfer. 
    \item Discussed computational cost and retrieval efficiency in context augmentation.
\end{itemize}

\section{Related Works}
Large language models, typically in the form of big neural networks, are trained with a huge amount of training data rich in unstructured knowledge.
Researchers have found that after model training, the neural networks often end up storing a surprisingly large amount of memorized information within its weights~\cite{arpit2017memorization, carlini2021memorization} which are then leveraged as a knowledge base.
Multiple hypotheses have been developed on how components such as fully-connected layers~\cite{geva2020transformer} and attention layers~\cite{bricken2021transformer} may be responsible for such memorization behavior. 
While the capability of storing world knowledge is desirable, memorization also contributes to huge model sizes and the lack of explicit control over knowledge base, such as performing selection or updates.

An alternative strategy is to enable language models to incorporate world knowledge in the form of retrieved context from external sources, instead of having to memorize them. 
Multiple works have proposed architectures that support external retrieval, usually composed of a context retriever that searches a large external key value store and a method of integrating retrieved information.

There are various ways to construct the key value store. The keys are primarily used for similarity matching, and they can be sparse vectors such as BM25~\cite{robertson2009probabilistic}, or dense embeddings extracted from part of the model~\cite{khandelwal2019knnlm, lewis2020pre,wu2022memorizing}, or from pretrained embedders~\cite{devlin2018bert, ni2021sentencet5}, or embedders trained for specific downstream tasks~\cite{guu2020realm, vladimir2020emnlp,xiong2020approximate,seo2019real,lee2020learning,de2021mention}. The values also have various different forms. For example, TOME~\cite{de2021mention} stores a dense embedding about the contextual information of an entity mention, while Realm~\cite{guu2020realm}, RAG~\cite{lewis2020rag}, FID~\cite{izacard2020fid}, Retro~\cite{borgeaud2021retro}, MARGE~\cite{lewis2020pre}, DenSPI~\cite{seo2019real} and DensePhrases~\cite{lee2020learning} store the raw text as the value. Works such as $k$NN-LM~\cite{khandelwal2019knnlm} and Spalm~\cite{yogatama2021spalm} store one token as a value. Finally the key value store is searched over using vector similarity techniques, typically with some off-the-shelf nearest neighbor search implementations such as FAISS~\cite{johnson2019faiss}, ScaNN~\cite{guo2020icml}, HNSW~\cite{malkov2018efficient} or SPTAG~\cite{sptag2018}.

There are also many different ways to integrate the retrieved context. A popular approach is to concatenate the retrieval results with the original input and jointly process them. It has been adopted by works such as Realm~\cite{guu2020realm}, RAG~\cite{lewis2020rag}, and FiD~\cite{izacard2020fid}. Other works utilize some forms of cross attention for the context integration, such as the Chunked-Cross-Attention with input conditioning in Retro~\cite{borgeaud2021retro}, $k$NN Attention in Memorizing Transformer~\cite{wu2022memorizing}, Memory Attention in TOME~\cite{de2021mention} and Cross-Attention in MARGE~\cite{lewis2020pre}. For token level integration, $k$NN-LM~\cite{khandelwal2019knnlm} uses simple linear interpolation while Spalm~\cite{yogatama2021spalm} uses a learned gate based on the last layer embedding. There are also works that directly utilize the retrieval results without any integration, such as DenSPI~\cite{seo2019real} and DensePhrases~\cite{lee2020learning}. Most of the works use retrieval as a way to augment tasks such as language modeling or question answering, with the exception of MARGE~\cite{lewis2020pre} where retrieval is treated as an autoencoder bottleneck for multilingual pretraining, and is not strictly necessary for inference. We compare representative previous works and contrast with our proposal in Table~\ref{tab:related}.

\begin{table}[t]
\begin{center}
\small
\bgroup
\def\arraystretch{1.1}
\begin{tabular}{|c||c|c||c|c||c|}
\hline
\shortstack{\\ Method \\ \vspace{.02 in}} &
\shortstack{Retrieval\\ Granularity} & \shortstack{Retrieval\\ Encoding} &
\shortstack{Context\\ Integration} & \shortstack{\\Decoupled\\ Context Encoding} 
& \shortstack{\\ Tasks \\ \vspace{.02 in}} \\
\hline
\hline
$k$NN-LM~\cite{khandelwal2019knnlm} & Token & Last layer & Interpolation & Yes & LM \\
\hline
Spalm~\cite{yogatama2021spalm} & Token & Last layer & Gating & Yes & LM \\
\hline
\shortstack{\\Realm~\cite{guu2020realm}, RAG~\cite{lewis2020rag}, \\FID~\cite{izacard2020fid}} & 
\shortstack{\\ Input \\ \vspace{.01 in}}  & \shortstack{\\ Trained \\ \vspace{.02 in}}  & \shortstack{\\ Concat \\ \vspace{.02 in}} &  \shortstack{\\ No \\ \vspace{.02 in}}  & \shortstack{\\ OpenQA \\ \vspace{.01 in}} \\
\hline
\shortstack{\\ Retro~\cite{borgeaud2021retro}  \\ \vspace{.01 in}}& \shortstack{\\ Chunk \\ \vspace{.02 in}}  & \shortstack{\\ Frozen \\ \vspace{.02 in}}  & \shortstack{Chunked-\\Cross-Attention} & \shortstack{\\ No \\ \vspace{.01 in}} & \shortstack{\\LM, \\OpenQA} \\
\hline
 \shortstack{Proposed \\ \vspace{.05 in}} &  \shortstack{Chunk \\ \vspace{.06 in}}   & \shortstack{Frozen /\\ trained \\ \vspace{.03 in}} & \shortstack{\\Encoder-\\Decoder \\ Cross-Attention} &  \shortstack{\\ Yes \\ \vspace{.05 in}}  & \shortstack{\\LM, \\OpenQA \\ \vspace{.01 in}} \\
\hline 
\end{tabular}
\egroup
\vspace{.1 in}
\caption{Architectural differences between previous retrieval augmented model and ours in (i) context retrieval, (ii) context integration and (iii) targeted applications.}\label{tab:related}
\end{center}
\end{table}
\begin{figure*}
\begin{center}
\vspace{-.2in}
\includegraphics[width=0.88 \textwidth]{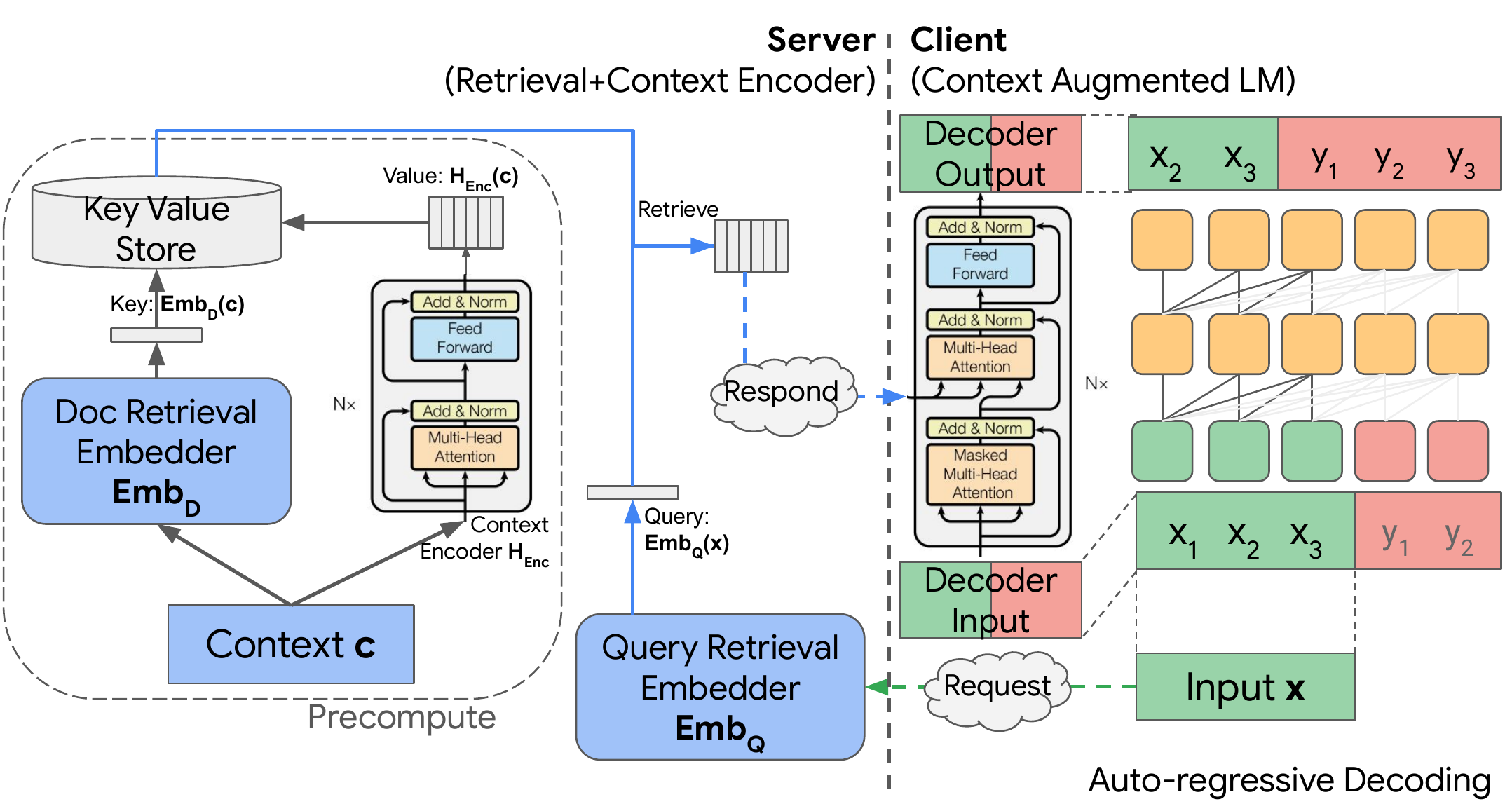}
\end{center}
\caption{\textbf{Architecture of context augmented language model.} We opt to use the standard \texttt{Encoder-Decoder} cross attention mechanism for context incorporation which allows us to decouple context encoding from LM inference. $\c$, $\x$, $\y$ serve as context, decoder input and decoder target respectively. In other words, the client sends over input $\x$, and the server conducts retrieval to find relevant context and returns the encoded representation $\Henc(\c; \mathbf{\theta}_{\texttt{Enc}})$. The encoded representation $\Henc$ is pre-computed offline and returned as ``metadata'' of the retrieval. Note that training of the \texttt{Encoder} and \texttt{Decoder} is joint, while they are decoupled at inference time: the client does not need to store parameters or the run inference on the \texttt{Encoder} component.}\label{fig:intuition}
\end{figure*}

\section{Architecture}
We use \texttt{Encoder-Decoder} Transformer architecture~\cite{vaswani2017attention} to integrate language model input and retrieved context. We denote the context encoder and LM decoder as \texttt{Enc} and \texttt{Dec} respectively. Given an input token sequence $\x = (x_1, x_2, ..., x_n)$, the task is to predict the next tokens $\y=(y_1, y_2, ..., y_s)$. 

Without external context, we concatenate $\x$ before $\y$ and the task becomes a traditional auto-regressive language modeling to predicts targets $\y$ following input (or ``promopt'') $\x$. 
In this setting, only the decoder is involved (denoted as ``No-retrieval’’). 
To incorporate external context, we use $\c$, $\x$, $\y$ to serve as encoder input, decoder input and decoder target respectively. We first use a retriever to identify the context $\c=(c_1, c_2, …, c_m)$ given input $\x$, then fetch the the pre-computed encoder output of the corresponding context tokens $\Henc(\c; \theta_{\texttt{Enc}})$ as output. 
In this setting, encoder output $\Henc$ is directly used by the decoder through Encoder-Decoder cross-attention to influence the final prediction. The decoder does not have to know the exact tokens of $\c$ that are retrieved.
\begin{gather*}
\texttt{No-retrieval:} P(y_i|y_{<i}, x_1, x_2, … x_n; \theta'_{\texttt{Dec}}) \\
\texttt{Retrieval:} P(y_i|y_{<i}, x_1, x_2, … x_n, \{\Henc(\c)\}; \theta_{\texttt{Dec}})
\end{gather*}
Under this formulation, only decoder parameters $\mathbf{\theta}_{\texttt{Dec}}$ are required at inference time.
The retriever retrieves indices of the relevant context and looks up their encodings. 
The context encodings are processed ahead of the time, and are completely decoupled from online operations.
This is in contrast to previous works of Realm~\cite{guu2020realm}, Rag~\cite{lewis2020rag} or Retro~\cite{borgeaud2021retro} where the interaction between input $\x$ and context $\c$ is bi-directional, which necessitates context encoding at inference time.
In our model, information flows uni-directionally from $\c$ to $\x$ and $\y$, and that the encoding of each context $\c$ is processed independently.
On one hand, this is more restrictive than bi-directional interaction;
on the other hand, such a design ensures complete decoupling of context processing and the online language model inference. The exact mechanism is detailed in Figure.~\ref{fig:intuition}.

Conceptually the retriever can be an arbitrary blackbox.
In practice, we use a dual encoder formulation~\cite{cer2018universal, henderson2017efficient}, which first represents $\x$ as a query embedding $Emb_Q(\x)$ and performs vector similarity search over a database of $\mathcal{D}$ to find the indices of documents whose document embedding has the highest inner products with the query embedding.
We then look up the context encoder outputs that correspond to retrieved indices and return them as the retriever output. 
\begin{gather*}
l^* = \argmax_{ \mathbf{v} \in \mathcal{D}} \langle Emb_Q(\x), \mathbf{v}) \rangle; \mathcal{D}=\{Emb_D(\c); \c\in \mathcal{C}\}\\
\Henc[l^*]= \texttt{Enc}(\c_{l^*}; \theta_{\texttt{Enc}})
\end{gather*}

In the case of multiple supporting context, $\kargmax$ is used instead of $\argmax$. The encoder outputs of each supporting context are then concatenated:
\begin{gather*}
P(y_i|y_{<i}, x_1, x_2, … x_n, Concat(\Henc[l_1], \Henc[l_2], \cdots \Henc[l_k]); \theta_{\texttt{Dec}});
\end{gather*}
Where $Concat$ is simply vector concatenation:
\begin{gather*}
Concat((\mathbf{h}_1, \mathbf{h}_2, \cdots, \mathbf{h}_n], [\mathbf{g}_1, \mathbf{g}_2, ..., \mathbf{g}_m], ...) = [\mathbf{h}_1, \mathbf{h}_2, ..., \mathbf{h}_n, \mathbf{g}_1, \mathbf{g}_2, ..., \textbf{g}_m, ...]
\end{gather*}

At training time, the encoder and decoder are jointly trained. We first perform offline retrieval to form triplets of $(\x, \y, \c)$, where $\c$ is retrieved by some predefined retriever. The loss is masked and only defined on the targets $\y$.
Because the encoding of each context is independent and there is no interaction between context, the attention matrix of encoder is block diagonal and we process them in a linear loop over each diagonal block.
Thus, the computation cost of both encoder and decoder at each step is linear in the number of context.

For online language model inference, only the retriever and the decoder are involved.
The retrieval embedding $Emb_{D}(\c)$ and the encoder output $H_{\texttt{Enc}\mathbf(c)}$ of the context are both offline pre-processed and prepared into the retrieval database of such associated key-value pairs.
When a new input sequence arrives, the retriever is only responsible for the approximate nearest neighbor search and lookup of the associated value that is the pre-computed encoder output.
The decoder then takes in the input sequence and cross attends on the concatenation of pre-computed encoder output to generate the targeted tokens.
Thanks to the decoupling, neither the retriever nor the decoder needs to store encoder parameters.
Hence, such an approach is more parameter efficient compared to similar works such as Retro~\cite{borgeaud2021retro} by saving the storage and computation budget on encoder, which is helpful in ``client-server'' scenario where the capacity of the ``client'' can be limited.
When accounting the parameter count in comparison with other models, we only need to count the decoder and cross attention parameters. 
We also followed Retro's~\cite{borgeaud2021retro} approach of excluding the embedding matrices from the parameter count. 

\section{Auto-regression Language Modeling}
\label{sec:lm-experiment}

We experimented with the same ``encoder-decoder'' context incorporation mechanism for both auto-regressive language modeling and open domain question answering. The only difference is that auto-regressive language modeling processes input sequences in a sliding window fashion, while question answering task receives the full input sequence (the question) at once.

\subsection{Experimental Setup}

\begin{figure*}
\hspace{-0.25 in}
\begin{subfigure}[b]{0.55 \textwidth}
\includegraphics[width=\textwidth]{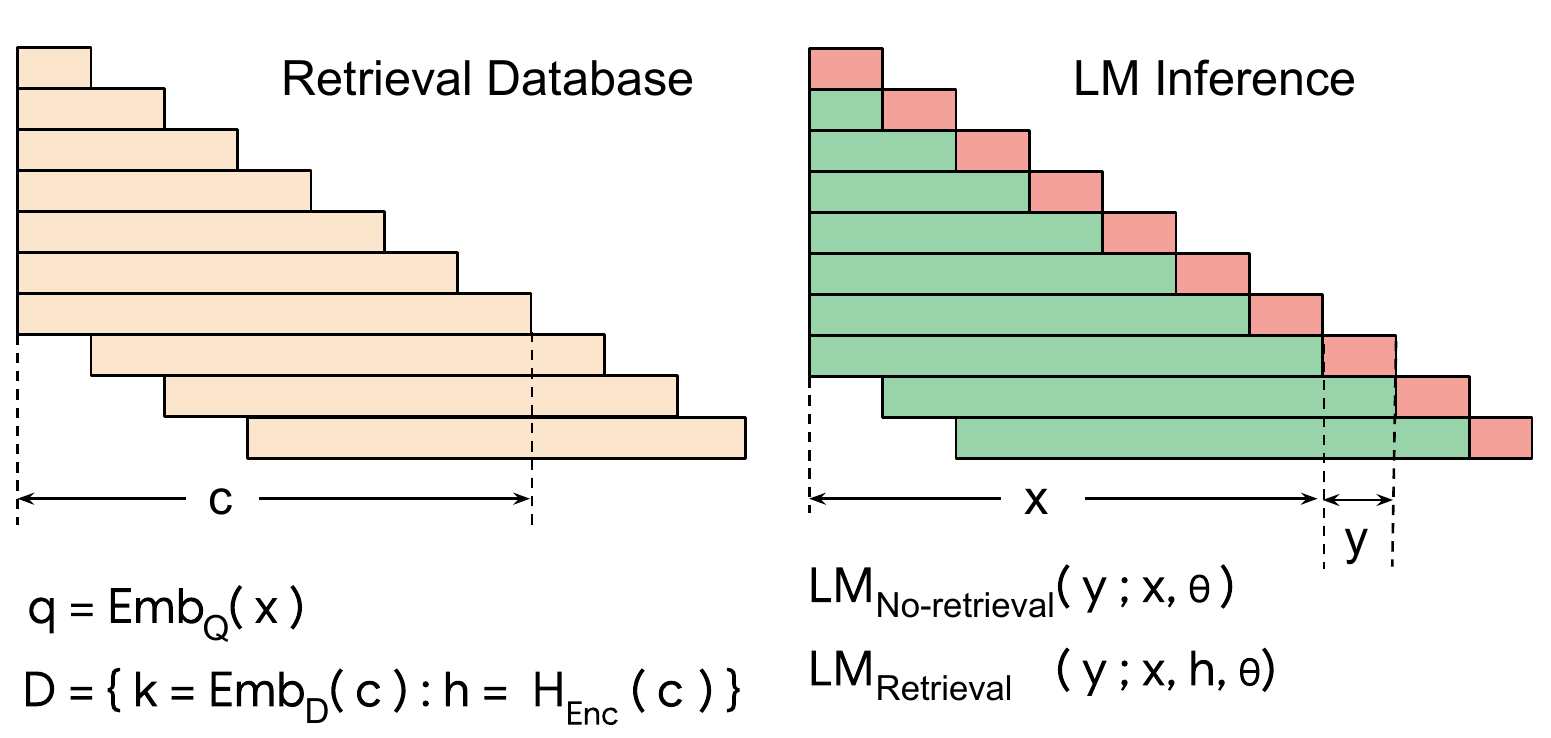}
\vspace{-0.2in}
\caption{}\label{fig:sliding}
\end{subfigure}
\begin{subfigure}[b]{0.4 \textwidth}
\small
\begin{tabular}{|c||c|c|}
\hline
Database Split & \# Articles & \# Entries \\
\hline
\hline
\texttt{C4 Train} & 364.6M & 3382M \\
\texttt{C4 Val, unfiltered} & 0.3645M & 3.369M \\
\texttt{C4 Val, filtered} & - & 2.868M \\
\hline
\texttt{NQ-open Train} & 79k & - \\
\texttt{NQ-open Dev} & 8.8k & - \\
\texttt{NQ-open Test} & 3.6k & - \\
\texttt{Wiki Database} & - & 21M \\
\hline
\end{tabular}
\vspace{0.05 in}
\caption{}\label{tab:entries}
\end{subfigure}
\caption{(a) Chunking scheme for language modeling training with C4. For each article in C4, we divide the tokenized text into non-overlapping blocks of at most 64 tokens and use them as targets $\y$. We use the preceding tokens of at most 448 in length as $\x$. (b) The number of entries in the retrieval database of C4 auto-regressive LM and Natural Question QA task. }\label{fig:database}
\end{figure*}

For auto-regressive language modeling, we use English C4~\cite{2020t5} version \texttt{2.2.1}, the same as Retro. We train the language model and prepare the retrieval database using the \texttt{train} split and evaluate the results using \texttt{validation} split. The language model target sequence is a sliding window (chunk) of $s=64$ tokens, with at most $n=448$ preceding tokens are used as input sequence. This setup is similar to XLNet~\cite{yang2019xlnet} and Retro~\cite{borgeaud2021retro}. The target and input sequences that are smaller than the given window size (64 and 448, respectively) are padded with zeros.

To construct the retrieval database, the same sliding window processing is also used for the context sequences. The database is formed as associated pairs of retrieval embedding and encoder output: $\{Emb_{D}(\mathbf{c}) : H_{\texttt{Enc}}(\mathbf{c})\}$, where context $\mathbf{c}$ are the sliding window of 512 tokens with a stride of 64 tokens. 
We choose our hyper-parameters to be comparable to Retro: chunk size $s=64$ and input window size $n=448$ (smaller than $2048$ that of Retro). This also implies that the number of entries in the database is larger than the number of articles, but smaller than the number of tokens. 

Our training corpus is in the form of triplets $(\x, \y, \mathbf{c})$. $\x$ and $\y$ are acquired directly by applying sliding window on the \texttt{train} split of C4. 
Then BM25~\cite{robertson2009probabilistic} is used as a bootstrapping retriever to mine relevant context $\mathbf{c}$ to from the database. 
The first retrieval results with no more than 8 consecutive token overlap with the target is used as context.
Figure~\ref{fig:sliding} illustrate the sliding window construction of database as well as sequence served as input and targets. Table~\ref{tab:entries} gives the exact number of entries in the resulting database used as external context.

We use mT5~\cite{xue-etal-2021-mt5} as the backbone architecture for our context augmented \texttt{Encoder-Decoder}, and train our models from scratch. \texttt{Train} split is used both for training and retrieval, while \texttt{validation} split is used for bits-per-byte (Bpb) evaluation.
In auto-regressive language model evaluation, due to the fact that text are crawled from web sources, there can be a non-trivial overlap of tokens between the training and validation splits. 
In such cases, tokens are often ``copy-pasted'' from retrieved context into targets without changing.
Such copying leads to near-zero perplexity on targets and has a big effect on final bits-per-byte measurement. 
Following the discussion on dataset leakage of Retro~\cite{borgeaud2021retro}, we filtered any example whose targets and context sequences have more than 8 common consecutive tokens (correspond to 12.5\% filtering of Retro). We found $14.87\%$ of the validation chunks are removed by the filtering of longest common substring.

\begin{figure*}%
\begin{subfigure}[b]{0.5 \textwidth}
\hspace{-0.5in}
\includegraphics[width=1.2 \textwidth]{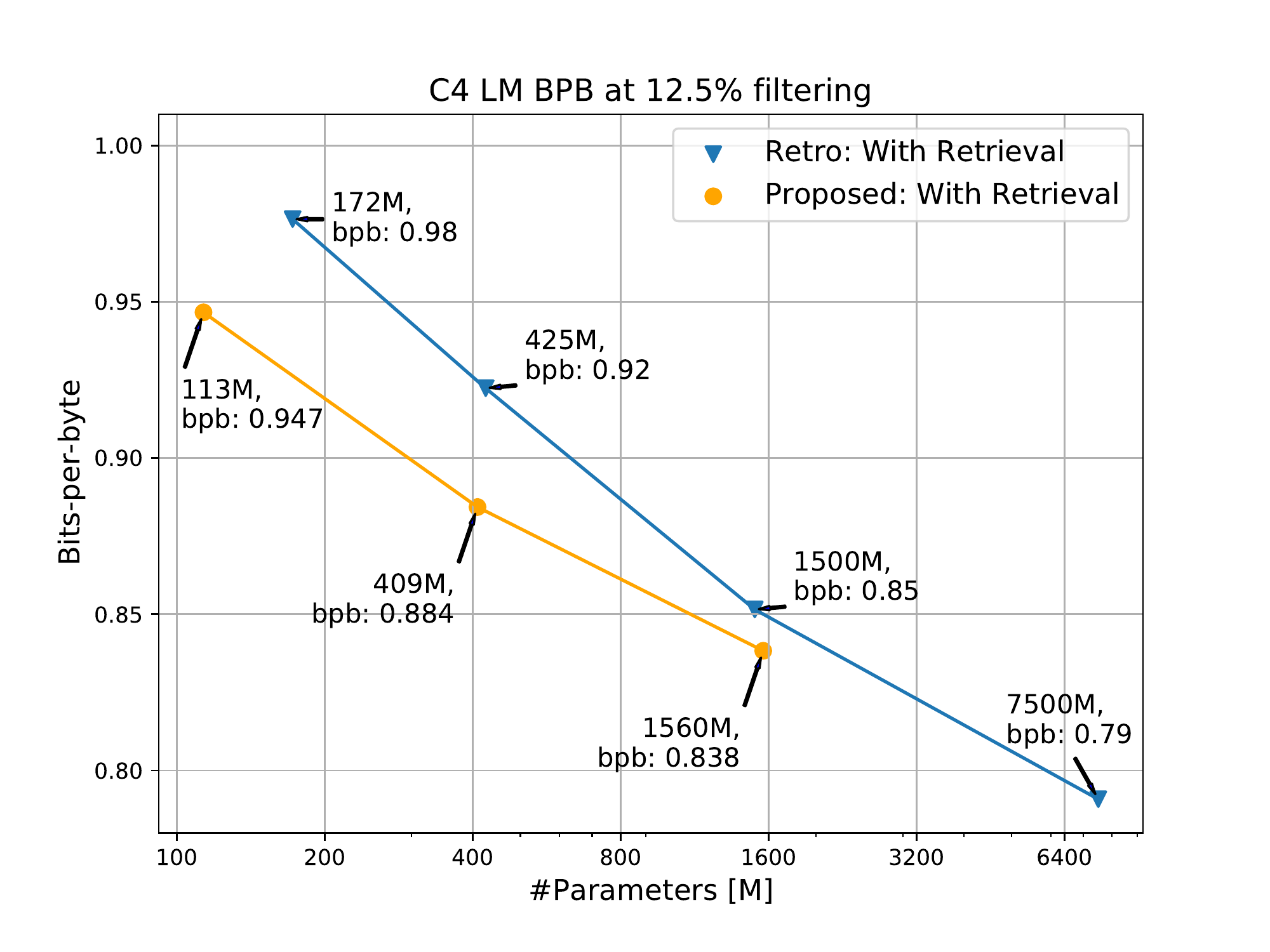}
\caption{}\label{fig:vs_retro}
\end{subfigure}
\begin{subfigure}[b]{0.40 \textwidth}

\bgroup
\def\arraystretch{1.1}
\begin{tabular}{|l|c|r|}
\hline 
Model & \shortstack{\\Model\\ Size} & \shortstack{\\Exact Match\\ Accuracy} \\
\hline 
\texttt{Realm}~\cite{guu2020realm} & 110M & 40.4 \\ 
\texttt{DPR}~\cite{vladimir2020emnlp} & 110M & 41.5 \\  
\texttt{\textbf{Ours (Large)}} & 409M & 44.35 \\
\texttt{RAG}~\cite{lewis2020rag} & 400M & 44.5 \\
\texttt{Retro}~\cite{borgeaud2021retro} & 7.5B & 45.5 \\
\texttt{\textbf{Ours (XL)}} & 1.56B & 47.95 \\
\hline 
\texttt{FiD}~\cite{izacard2020fid} & 770M & 51.4 \\    
\texttt{EMDR}~\cite{singh2021end} & 440M & 52.5 \\
\texttt{FiD + Distill}~\cite{izacard2020distilling} & 770M & 54.4 \\
\hline 
\end{tabular}
\egroup
\vspace{0.3in}
\caption{}\label{tab:nq_result}
\end{subfigure}
\caption{ Performance of decoupled encoder-decoder on auto-regressive language modeling (C4) and question answering tasks (Natural Question). (a) Comparing the proposed method and Retro on \texttt{c4-en-2.2.1} validation split with 12.5\% token overlap filtering. The y-axis measures bits-per-byte (bpb, lower is better), which is the perplexity normalized by token length. The x-axis shows the number of non-embedding parameters in log scale. (b) End-to-end result on Natural Question test split. Following previous works, we measure exact match (EM) accuracy on ``short answer type'' with at most five tokens.
\vspace{-0.2in}
}\label{fig:results}
\end{figure*}

Figure~\ref{fig:vs_retro} reports our auto-regressive language model results with different model sizes under the Bit-per-byte (bpb) values. Bpb is tokenizer agnostic and is often used to compare models with different vocabulary.
We use mT5 \texttt{base}, \texttt{large} and \texttt{XL} respectively without modification, and the results compare favorably to Retro models with similar sizes that uses customized chunked-cross-attention which couples context encoding and LM decoding. 
Our experiments demonstrate that context incorporation can be achieved with simple \texttt{Encoder-Decoder} cross attention, with the additional benefit of decoupled encoder processing. 

\subsection{Retriever Training}
The goal of context retrieval is to identify the context sequences $\c^*$ that maximize the improvement of some utility function, such as the log-likelihood improvement on target prediction. i.e.

\begin{gather*}
\c^* = \argmax_{\c} U(\x, \y, \c); \\
U(\x, \y, \c) = \sum_{y_i} \log P(y_i|y_{<i}, \x, \c, \mathbf{\theta}) - \log P(y_i|y_{<i}, \x, \mathbf{\theta'})
\end{gather*}

However, it is infeasible to evaluate the utility function $U(\x, \y, \mathbf{c})$ for all triplets $(\x, \y, \mathbf{c})$. Therefore, we introduce a proxy to approximate $U$ by computing the expectation of $U$ conditioned on the token $y_i$ and whether it appeared in the input $\x$ and context sequence $\c$:
\begin{gather*}
\hat{U}(\x, \y, \mathbf{c}) = \sum_{y_i\in \y, \mathbf{c}} \mathbbm{1}_{y_i \in \x}  \bar{U}(y_i|y_i \in \x, y_i \in \mathbf{c}) + \mathbbm{1}_{y_i \notin \x}  \bar{U}(y_i|y_i \notin \x, y_i \in \mathbf{c})
\end{gather*}
Intuitively, the ``context utility'' $\hat{U}$ is a weighted token overlap between context $\c$ and target $\y$.
The weight is higher if the token $y_i$ did not appear in input but is contained in context, and when the token is ``sensitive'' to context by showing larger loglikelihood change when the context is present.
To train a retriever, we adopted the typical dual encoder formulation with in-batch softmax training similar to that of DPR~\cite{vladimir2020emnlp, henderson2017efficient}.
We use training data bootstrapped from BM25 retrieval, where (1) the valid retrieval with highest context utility $\hat{U}$ is the positive; (2) the other top retrieval from BM25 that has less than 80\% of context utility of the highest one is used as hard negative; (3) The rest of the in-batch samples are regarded as random negatives. All retrievals are subject to the filtering criteria of no more than 8 consecutive tokens.

We trained our retriever using T5X retrieval framework~\cite{ni2021sentencet5} based on mT5-\texttt{small} with an embedding dimensionality of 128.
The model is trained for 100,000 iterations on a batch size of 8,192 on 64 TPUv3 chips.
Unfortunately, using the dense trained retriever does not lead to a visible improvement on the final C4 Bpb evaluation, possibly because both the dense and BM25 retriever end up retrieving similar context. We report results using trained dense retriever because of the better retrieval efficiency.

\subsection{Ablation and Analysis} \label{sec:long-short} 
\begin{figure*}%
\vspace{-.25in}
\hspace{-.25in}
\begin{subfigure}[b]{0.5 \textwidth}
\includegraphics[width=1 \textwidth]{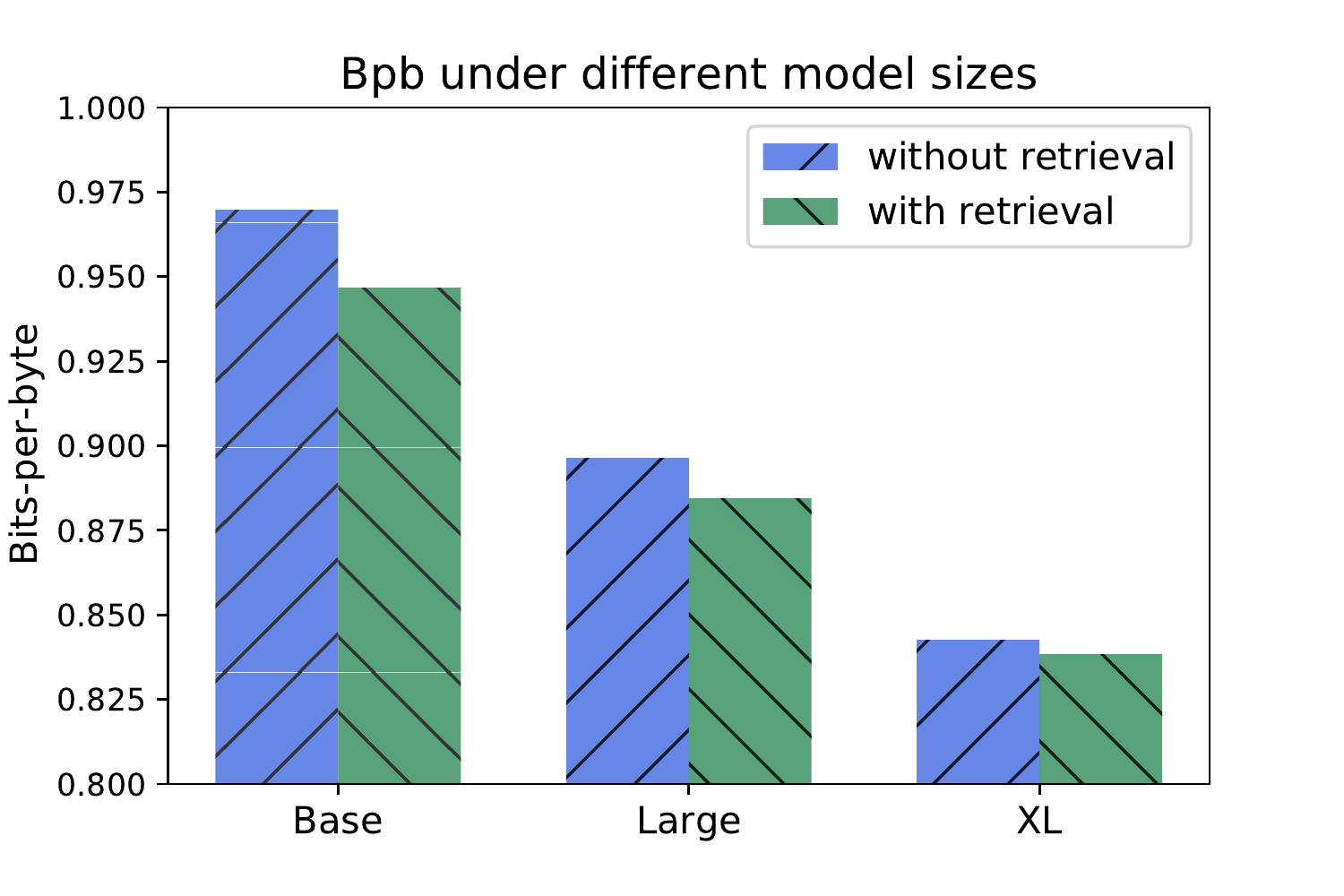}
\vspace{-.25in}
\caption{}\label{fig:c4_ablation}
\end{subfigure}
\begin{subfigure}[b]{0.5 \textwidth}
\includegraphics[width=1.1 \textwidth]{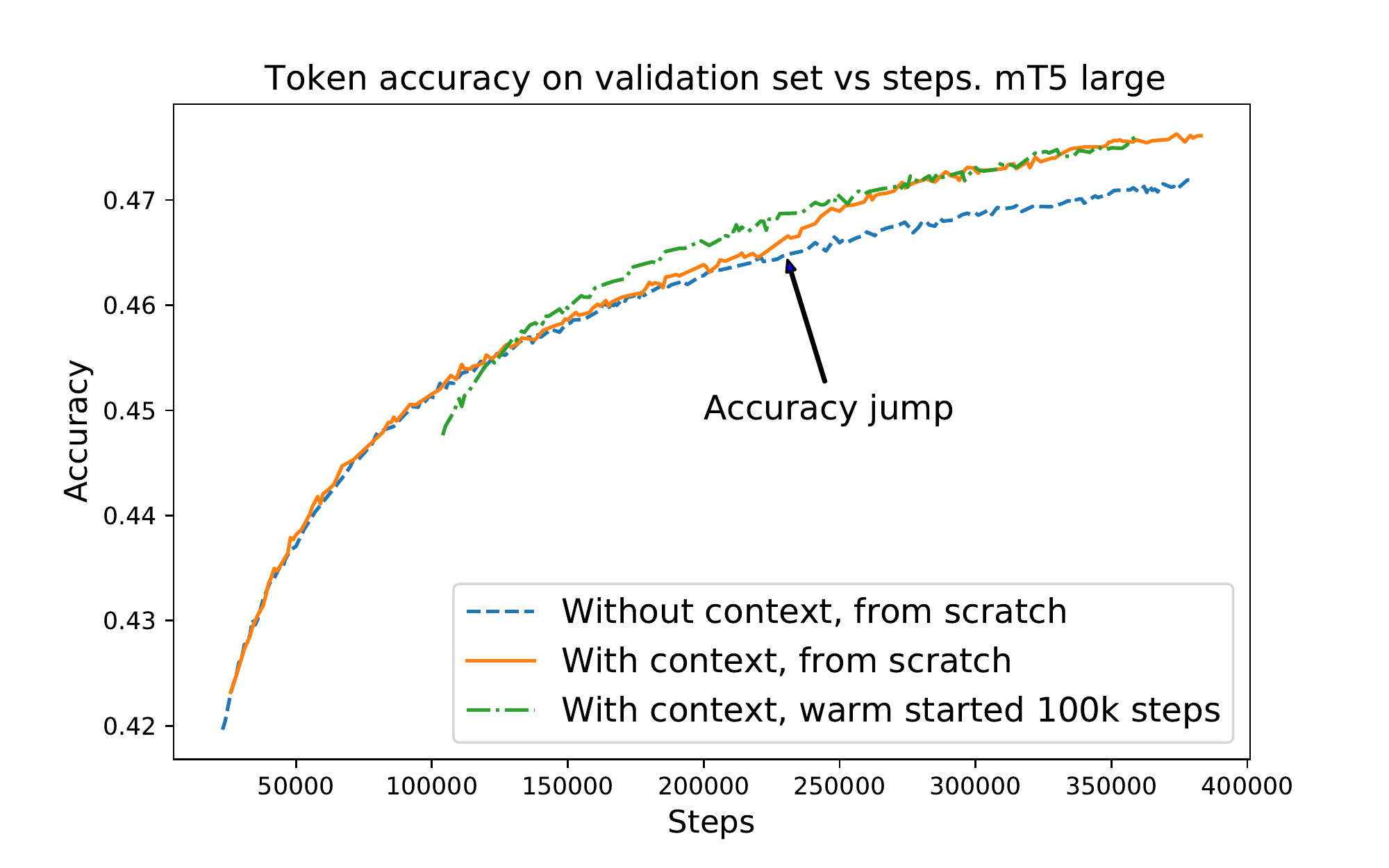}
\vspace{-.25in}
\caption{}\label{fig:jump_warmstart}
\end{subfigure}
\caption{(a) Bpb evaluation (lower is better) comparing models of the same sizes trained from scratch, with and without context retrieval, respectively. (b) Comparing the accuracy history of \texttt{Large} model training with and without retrieval. Notice that the curve ``jumps'' when training with retrieval but without warm-starting.}
\end{figure*}

To analyze the effect of the context augmentation, we show first in Figure~\ref{fig:c4_ablation} that the context augmentation in language model consistently helps across different model sizes in the proposed method. 
While the absolute improvement on Bpb over ``No-retrieval'' may seem small when it is filtered to 12.5\% in longest common sequence, we note that the improvements were made on a limited set of important tokens.
Intuitively, tokens of functional words such as ``the'', ``a'', ``to'' should not benefit much from external information and can be predicted with its local context. 
Yet functional words make up a significant portion in perplexity because of their high frequencies in text.

\begin{figure*}%
\vspace{-.1in}
\begin{center}
\includegraphics[width=1 \textwidth]{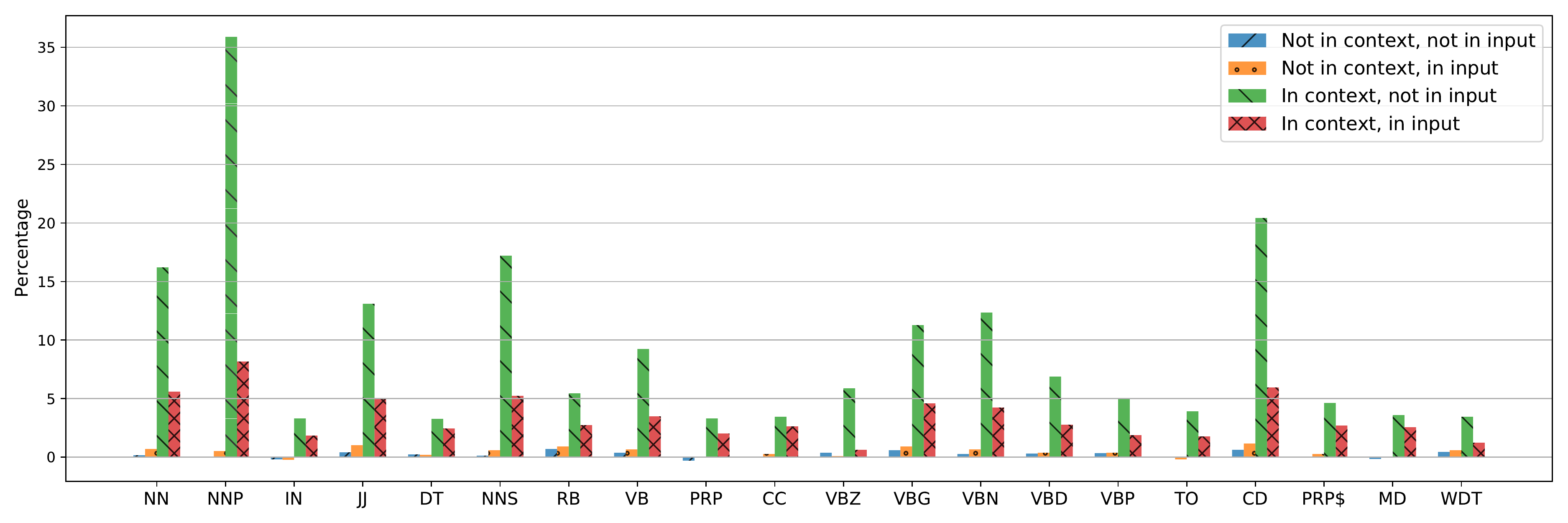}
\end{center}
\caption{Percentage of improvement on target log-likelihood when there's context. We show the most frequent 20 POS tags, excluding punctuation. For the data on all the POS tags, see Appendix~\ref{tab:pos_percent}.
}\label{fig:in_long_in_short}
\end{figure*}

To this end, we consider the difference in log-likelihood of two language models of same size, trained with and without retrieved context.
We evaluate the models on validation split of C4 and obtain a breakdown using part-of-speech (POS) tagging.
We used \texttt{NLTK}~\cite{bird2009natural} to assign each target token to a POS tag. Note that the tokenization boundaries of \texttt{NLTK} and sentencepiece~\cite{kudo2018sentencepiece} could be different, and the assignment is done by majority voting in that case. 

From Figure~\ref{fig:in_long_in_short}, it is immediately clear that the impact of context retrieval on different types of tokens is not uniform.
Noun (\texttt{NN}, \texttt{NNP}, \texttt{NNS}, \texttt{NNPS}), and number (\texttt{CD}) benefited the most from the context, followed by adjective (\texttt{JJ}) and verb (\texttt{VB}, \texttt{VBZ}, \texttt{VBG}, \texttt{VBN}, \texttt{VBD}, \texttt{VBP}). 
Tokens such as preposition (\texttt{IN}), coordinating conjunction (\texttt{CC}), etc. are less helped by context.
We believe it is because their prediction usually only depends on immediate local context.
We also found that whether the target token appeared in the context sequence or input sequence has a large impact on the improvements, and included this in the breakdown.
Intuitively, tokens that are grounded in the retrieved context $\mathbf{c}$ but did not appear in the input $\x$ benefit the most from context retrieval.

\subsubsection{Training Details}
We trained the Encoder-Decoder LM model for a total of $1,100,000$ steps with a batch size of 512 and a default learning rate schedule of square-root decay.
This corresponds to $10,000$ warmup steps with a fixed learning rate of $0.01$, followed by square-root decay for $990,000$ steps. 
We found training from scratch and fine-tuning on pre-trained models result in similar performance on C4. Therefore we opted to train from scratch. We also found it beneficial to include additional $100,000$ steps with a lower learning rate of $2 \times 10^{-4}$ after standard training schedule, which slightly improves the results.
For \texttt{base} and \texttt{large} we used 64 TPUv3 chips whereas 128 TPUv3 chips for training \texttt{XL}. \texttt{XL} runs 1.5 training steps per second. \texttt{XXL} is left out due to the insufficient computation resources.

Comparing the training curve of model with and without retrieval, as depicted in Figure~\ref{fig:jump_warmstart}, we noticed a phenomenon where both models have roughly the same performance initially but the one with retrieval suddenly increases at some point and continues to improve afterwards.
This is especially true in \texttt{large} and \texttt{XL}. We suspect that decoders of larger models have higher memorization capacity, and in the initial phase they can improve training objective without delegating memorization to context retrieval. 
This is a challenge to training stability and one way to address this is by first ``warm start'' the model with $100,000$ steps of training on a subset of training data with highest 10\% of the context utility before proceed on the full dataset.
\section{Open Domain Question Answering}
\label{sec:nq}
\subsection{Natural Question} \label{sec:natural_question}
Large language models are useful because they generalize to downstream tasks, in addition to performing auto-regressive language generation.
To demonstrate that the proposed context augmentation scheme is effective on downstream applications we evaluated our proposed model of decoupled \texttt{Encoder-Decoder} on the OpenQA task of Natural Question~\cite{naturalquestion}.

We use the same question and context processed by~\cite{izacard2020fid}, where the context is retrieved with DPR retriever~\cite{vladimir2020emnlp}.
We construct each context sequence in the format of \texttt{"title:~\{title\}~source:~\{source\}"} and pad or trim it to 256 sentencepiece tokens.
The context sequences are then encoded by the encoder independently and simply concatenated, to ensure there is no interaction between them.

The decoder is trained to receive and respond on the sequence containing both the question and answer in format \texttt{"question:~\{question\}~\textbackslash n~answer: \underline{\{answer\}}"} similar to~\cite{borgeaud2021retro}, where \texttt{"question:~\{question\}~\textbackslash n~answer:~"} is the input prompt (and there is no loss defined on these tokens) and \texttt{\underline{\{answer\}}} is the target sequence the decoder is expected to predict. Again, the decoder cross-attends to the concatenated encoder output $\Henc(\c)$ from all the retrieved context and the decoder is completely decoupled from encoder. It does not even have to know the exact tokens of context sequence being returned by the retriever.

The model weights are initialized from T5.1.1 checkpoints~\cite{2020t5}
We jointly fine-tuned the encoder and decoder for $40,000$ steps with 20 context passages for each input of the \texttt{train} split, and validated on the \texttt{dev} split every 1,000 steps.
We used a batch size of $64$, a fixed learning rate of $10^{-4}$ and Adafactor optimizer~\cite{shazeer2018adafactor}. 
Finally, we selected the checkpoint with the best validation accuracy on the \texttt{dev} split and evaluated on the \texttt{test} split.
Each model is trained on 64 TPUv3 chips, and the evaluation metrics is string exact match with SQuAD~\cite{rajpurkar2016squad} normalization.
We report our final results in Table~\ref{tab:nq_result}.
Overall, the decoupled \texttt{Encoder-Decoder} produced competitive results on Natural Question task, albeit being a much simpler architecture and not specifically designed for QA tasks.
We note that FiD~\cite{izacard2020fid} uses similar architecture. However, FiD concatenates the question with the retrieved context passages. Therefore the context encoding is dependent on the input sequence and needs to be computed at inference time.

\subsection{Grounded Answer Generation}
Context augmented models tend to generate answers by transferring tokens from context to the output.
It is interesting to quantify how often the model output comes from grounded transfer of context tokens and how often this leads to the correct answer.
We use the same model trained in~\ref{sec:natural_question} to (a) run language model decoding with the original set of retrieved context to obtain ``original output'': $\y^*=Decode(\x, \C; \mathbf{\theta})$; and (b) rerun model prediction but remove all context passages that contain the ``original output'': $\y'=Decode(\x, \C'=\C \setminus \{\c; \y^* \subseteq \c\}; \mathbf{\theta})$.
In the case of grounded context transfer, we expect the ``secondshot output'' to change, but still grounded in the remaining context.

We show grounded-ness analysis in Table~\ref{tab:nq_analysis}. 
Overall, we found 76\% (2,765 out of 3,354) of the cases are likely due to grounded context transfer, because the ``secondshot output'' has changed to something different from ``original output'', but is still grounded in the remaining context passages.
In addition, we found that the accuracy is 51.5\% when model output is from grounded transfer, higher than 35.4\% for the rest. 
This implies that the model has higher accuracy when the output is grounded.
We exclude 37 examples from the analysis because they have less than 20 neighbors after removing original output.

\begin{table}
\large
\resizebox{\columnwidth}{!}{%
\bgroup
\def\arraystretch{1.3}
\begin{tabular}{|c||cccc|c|}
\hline
\multirow{2}{*}{
\shortstack{Original output \\ $\y^*=Decode(\x, \C; \mathbf{\theta})$}} &
\multicolumn{4}{c|}{Original output in context} & 
Original output out of context \\ 
\cline{2-6} & 
\multicolumn{4}{c|}{3354} & 256 \\
\hline \hline
\multirow{2}{*}{\shortstack{\vspace{-.3in} \\ Secondshot output \\ $\y'=Decode(\x, \C'; \mathbf{\theta})$ \\ $\C'=\C \setminus \{\c; \y^* \subseteq \c\}$}} &
\multicolumn{1}{c|}{\shortstack{Output changed \\ still in context \\ $\y' \neq \y^*, \y' \subseteq \C'$}} &
\multicolumn{1}{c|}{\shortstack{Output changed \\ out of context \\ $\y' \neq \y^*, \y' \nsubseteq \C'$}} &
\multicolumn{1}{c|}{\shortstack{Output unchanged \\ \vspace{0.05in} }} &
\shortstack{Excluded \\ \vspace{0.08in} } & 
- \\ \cline{2-6} 
& \multicolumn{1}{c|}{2765} &
\multicolumn{1}{c|}{261} & 
\multicolumn{1}{c|}{291} & 
37 & - \\ \hline
\end{tabular}%
\egroup
}
\vspace{.1 in}
\caption{Analysis of grounded context transfer on Natural Question. First row shows the number of questions broken down by whether the original model output appeared in the retrieved passages. 
To demonstrate the model follows grounded context transfer, we removed context passages containing original output and reran inference on the remaining context passages. 76\% of output changed due to the removal of ``original output'' but the new prediction is still contained in remaining context.} \label{tab:nq_analysis}
\vspace{-.2 in}
\end{table}

\section{Computational Discussion}
\label{sec:computation}

We evaluate the computation implications of our method around inference latency and space-time trade off. We estimate the latency based on benchmarks and publicly available packages and performance data.
In the Natural Question experiment, we use the 1.56B XL model in Table~\ref{tab:nq_result}, whose encoder output have dimension of 2048 and the activation data type is \texttt{bfloat16}~\cite{ieee754}. 
When profiled on a single TPUv3 core, it takes 200ms to encode all 20 retrieved context passages each containing 256 tokens (with paddings). This is equivalent to 10ms per context.
Thanks to the decoupling of the context encoding, the context passages encoding is done offline, and simply looked up at inference time. 
Inferencing the DPR query embedder model takes around 8ms unbatched on one TPUv2 chip.  
We use ScaNN~\cite{guo2020icml} to perform approximate similarity search on the DPR embeddings with a dimensionality of $768$, which takes roughly 12ms to retrieve 20 neighbors with recall@20=0.97 on a single CPU core. 
For each passage, the uncompressed encoded embeddings have a total size of 1 MiB (\texttt{sizeof(bfloat16) * 2048 * 256}). 
When reading the encoding from an NVMe SSD with a read throughput of 660 MB/s, the lookup takes around 1.5ms and we use another 0.8ms to transmit of one encoding over network. End to end, it takes around 66ms (\texttt{(1.5 + 0.8) * 20 + 12 + 8}) to retrieve 20 encoded passages. 
This is about a third of time for running inference of the encoder, and everything except query embedder is conducted with generic hardware without accelerators which is cheaper and consumes less energy.

However, one limitation of this approach is the amount of disk space and network bandwidth needed to store and transmit all the pre-computed encodings, which scales with the dimensionality of encoder output.
We leave it for future research to reduce the encoder output size while maintaining similar quality, which may be possible through compression, dimensionality reduction, or adding projection layer to the encoder.

\section{Conclusion}
In this paper, we advocate the design of context augmented language model based on \texttt{Encoder-Decoder} architecture.
\texttt{Encoder-Decoder} models are simple, proven and enjoy the unique computational advantage that the context encoding and language generation are decoupled. 
That is, the context encoding can be produced with offline pre-computation and caching, while the decoder is parameter efficient because it is agnostic to the encoder computation. 
We further demonstrated the effectiveness of this simple architecture by comparing with competitive baselines on common NLP tasks such as language modeling on C4 and question answering on Natural Questions.
We also analyzed the model behavior and showed the context augmentation resulted in big improvements on content words and the model generates output grounded in retrieved context most of the time.
Finally, we estimated the latency improvement from caching enabled by the decoupled computation.

\bibliography{research}

\begin{thebibliography}{10}

\bibitem{ieee754}
Ieee standard for floating-point arithmetic.
\newblock {\em IEEE Std 754-2019 (Revision of IEEE 754-2008)}, pages 1--84,
  2019.

\bibitem{arpit2017memorization}
Devansh Arpit, Stanis{\l}aw Jastrz{\k{e}}bski, Nicolas Ballas, David Krueger,
  Emmanuel Bengio, Maxinder~S Kanwal, Tegan Maharaj, Asja Fischer, Aaron
  Courville, Yoshua Bengio, et~al.
\newblock A closer look at memorization in deep networks.
\newblock In {\em International conference on machine learning}, pages
  233--242. PMLR, 2017.

\bibitem{bird2009natural}
Steven Bird, Ewan Klein, and Edward Loper.
\newblock {\em Natural language processing with Python: analyzing text with the
  natural language toolkit}.
\newblock " O'Reilly Media, Inc.", 2009.

\bibitem{borgeaud2021retro}
Sebastian Borgeaud, Arthur Mensch, Jordan Hoffmann, Trevor Cai, Eliza
  Rutherford, Katie Millican, George van~den Driessche, Jean-Baptiste Lespiau,
  Bogdan Damoc, Aidan Clark, et~al.
\newblock Improving language models by retrieving from trillions of tokens.
\newblock {\em arXiv preprint arXiv:2112.04426}, 2021.

\bibitem{bricken2021transformer}
Trenton Bricken and Cengiz Pehlevan.
\newblock Attention approximates sparse distributed memory.
\newblock {\em Advances in Neural Information Processing Systems}, 34, 2021.

\bibitem{brown2020gpt3}
Tom Brown, Benjamin Mann, Nick Ryder, Melanie Subbiah, Jared~D Kaplan, Prafulla
  Dhariwal, Arvind Neelakantan, Pranav Shyam, Girish Sastry, Amanda Askell,
  et~al.
\newblock Language models are few-shot learners.
\newblock {\em Advances in neural information processing systems},
  33:1877--1901, 2020.

\bibitem{carlini2021memorization}
Nicholas Carlini, Florian Tramer, Eric Wallace, Matthew Jagielski, Ariel
  Herbert-Voss, Katherine Lee, Adam Roberts, Tom Brown, Dawn Song, Ulfar
  Erlingsson, et~al.
\newblock Extracting training data from large language models.
\newblock In {\em 30th USENIX Security Symposium (USENIX Security 21)}, pages
  2633--2650, 2021.

\bibitem{cer2018universal}
Daniel Cer, Yinfei Yang, Sheng-yi Kong, Nan Hua, Nicole Limtiaco, Rhomni~St
  John, Noah Constant, Mario Guajardo-Cespedes, Steve Yuan, Chris Tar, et~al.
\newblock Universal sentence encoder.
\newblock {\em arXiv preprint arXiv:1803.11175}, 2018.

\bibitem{sptag2018}
Qi~Chen, Haidong Wang, Mingqin Li, Gang Ren, Scarlett Li, Jeffery Zhu, Jason
  Li, Chuanjie Liu, Lintao Zhang, and Jingdong Wang.
\newblock {\em SPTAG: A library for fast approximate nearest neighbor search},
  2018.

\bibitem{chowdhery2022palm}
Aakanksha Chowdhery, Sharan Narang, Jacob Devlin, Maarten Bosma, Gaurav Mishra,
  Adam Roberts, Paul Barham, Hyung~Won Chung, Charles Sutton, Sebastian
  Gehrmann, et~al.
\newblock Palm: Scaling language modeling with pathways.
\newblock {\em arXiv preprint arXiv:2204.02311}, 2022.

\bibitem{de2021mention}
Michiel de~Jong, Yury Zemlyanskiy, Nicholas FitzGerald, Fei Sha, and William
  Cohen.
\newblock Mention memory: incorporating textual knowledge into transformers
  through entity mention attention.
\newblock {\em arXiv preprint arXiv:2110.06176}, 2021.

\bibitem{devlin2018bert}
Jacob Devlin, Ming-Wei Chang, Kenton Lee, and Kristina Toutanova.
\newblock Bert: Pre-training of deep bidirectional transformers for language
  understanding.
\newblock {\em arXiv preprint arXiv:1810.04805}, 2018.

\bibitem{geva2020transformer}
Mor Geva, Roei Schuster, Jonathan Berant, and Omer Levy.
\newblock Transformer feed-forward layers are key-value memories.
\newblock {\em arXiv preprint arXiv:2012.14913}, 2020.

\bibitem{guo2020icml}
Ruiqi Guo, Philip Sun, Erik Lindgren, Quan Geng, David Simcha, Felix Chern, and
  Sanjiv Kumar.
\newblock Accelerating large-scale inference with anisotropic vector
  quantization.
\newblock In {\em International Conference on Machine Learning}, pages
  3887--3896. PMLR, 2020.

\bibitem{guu2020realm}
Kelvin Guu, Kenton Lee, Zora Tung, Panupong Pasupat, and Ming-Wei Chang.
\newblock Realm: Retrieval-augmented language model pre-training.
\newblock {\em arXiv preprint arXiv:2002.08909}, 2020.

\bibitem{henderson2017efficient}
Matthew Henderson, Rami Al-Rfou, Brian Strope, Yun-Hsuan Sung, L{\'a}szl{\'o}
  Luk{\'a}cs, Ruiqi Guo, Sanjiv Kumar, Balint Miklos, and Ray Kurzweil.
\newblock Efficient natural language response suggestion for smart reply.
\newblock {\em arXiv preprint arXiv:1705.00652}, 2017.

\bibitem{izacard2020distilling}
Gautier Izacard and Edouard Grave.
\newblock Distilling knowledge from reader to retriever for question answering.
\newblock {\em arXiv preprint arXiv:2012.04584}, 2020.

\bibitem{izacard2020fid}
Gautier Izacard and Edouard Grave.
\newblock Leveraging passage retrieval with generative models for open domain
  question answering.
\newblock {\em arXiv preprint arXiv:2007.01282}, 2020.

\bibitem{johnson2019faiss}
Jeff Johnson, Matthijs Douze, and Herv{\'e} J{\'e}gou.
\newblock Billion-scale similarity search with gpus.
\newblock {\em IEEE Transactions on Big Data}, 2019.

\bibitem{vladimir2020emnlp}
Vladimir Karpukhin, Barlas Oguz, Sewon Min, Patrick S.~H. Lewis, Ledell Wu,
  Sergey Edunov, Danqi Chen, and Wen{-}tau Yih.
\newblock Dense passage retrieval for open-domain question answering.
\newblock In {\em Proceedings of the 2020 Conference on Empirical Methods in
  Natural Language Processing, {EMNLP} 2020, Online, November 16-20, 2020}.

\bibitem{khandelwal2019knnlm}
Urvashi Khandelwal, Omer Levy, Dan Jurafsky, Luke Zettlemoyer, and Mike Lewis.
\newblock Generalization through memorization: Nearest neighbor language
  models.
\newblock {\em ICLR 2020}, 2019.

\bibitem{kudo2018sentencepiece}
Taku Kudo and John Richardson.
\newblock Sentencepiece: A simple and language independent subword tokenizer
  and detokenizer for neural text processing.
\newblock {\em arXiv preprint arXiv:1808.06226}, 2018.

\bibitem{naturalquestion}
Tom Kwiatkowski, Jennimaria Palomaki, Olivia Redfield, Michael Collins, Ankur
  Parikh, Chris Alberti, Danielle Epstein, Illia Polosukhin, Matthew Kelcey,
  Jacob Devlin, Kenton Lee, Kristina~N. Toutanova, Llion Jones, Ming-Wei Chang,
  Andrew Dai, Jakob Uszkoreit, Quoc Le, and Slav Petrov.
\newblock Natural questions: a benchmark for question answering research.
\newblock {\em Transactions of the Association of Computational Linguistics},
  2019.

\bibitem{lee2020learning}
Jinhyuk Lee, Mujeen Sung, Jaewoo Kang, and Danqi Chen.
\newblock Learning dense representations of phrases at scale.
\newblock {\em arXiv preprint arXiv:2012.12624}, 2020.

\bibitem{lewis2020pre}
Mike Lewis, Marjan Ghazvininejad, Gargi Ghosh, Armen Aghajanyan, Sida Wang, and
  Luke Zettlemoyer.
\newblock Pre-training via paraphrasing.
\newblock {\em Advances in Neural Information Processing Systems},
  33:18470--18481, 2020.

\bibitem{lewis2020rag}
Patrick Lewis, Ethan Perez, Aleksandra Piktus, Fabio Petroni, Vladimir
  Karpukhin, Naman Goyal, Heinrich K{\"u}ttler, Mike Lewis, Wen-tau Yih, Tim
  Rockt{\"a}schel, et~al.
\newblock Retrieval-augmented generation for knowledge-intensive nlp tasks.
\newblock {\em Advances in Neural Information Processing Systems},
  33:9459--9474, 2020.

\bibitem{malkov2018efficient}
Yu~A Malkov and Dmitry~A Yashunin.
\newblock Efficient and robust approximate nearest neighbor search using
  hierarchical navigable small world graphs.
\newblock {\em IEEE transactions on pattern analysis and machine intelligence},
  42(4):824--836, 2018.

\bibitem{ni2021sentencet5}
Jianmo Ni, Gustavo~Hern{\'a}ndez {\'A}brego, Noah Constant, Ji~Ma, Keith~B
  Hall, Daniel Cer, and Yinfei Yang.
\newblock Sentence-t5: Scalable sentence encoders from pre-trained text-to-text
  models.
\newblock {\em arXiv preprint arXiv:2108.08877}, 2021.

\bibitem{patterson2021carbon}
David Patterson, Joseph Gonzalez, Quoc Le, Chen Liang, Lluis-Miquel Munguia,
  Daniel Rothchild, David So, Maud Texier, and Jeff Dean.
\newblock Carbon emissions and large neural network training.
\newblock {\em arXiv preprint arXiv:2104.10350}, 2021.

\bibitem{rae2021gopher}
Jack~W Rae, Sebastian Borgeaud, Trevor Cai, Katie Millican, Jordan Hoffmann,
  Francis Song, John Aslanides, Sarah Henderson, Roman Ring, Susannah Young,
  et~al.
\newblock Scaling language models: Methods, analysis \& insights from training
  gopher.
\newblock {\em arXiv preprint arXiv:2112.11446}, 2021.

\bibitem{raffel2019t5}
Colin Raffel, Noam Shazeer, Adam Roberts, Katherine Lee, Sharan Narang, Michael
  Matena, Yanqi Zhou, Wei Li, and Peter~J Liu.
\newblock Exploring the limits of transfer learning with a unified text-to-text
  transformer.
\newblock {\em arXiv preprint arXiv:1910.10683}, 2019.

\bibitem{2020t5}
Colin Raffel, Noam Shazeer, Adam Roberts, Katherine Lee, Sharan Narang, Michael
  Matena, Yanqi Zhou, Wei Li, and Peter~J. Liu.
\newblock Exploring the limits of transfer learning with a unified text-to-text
  transformer.
\newblock {\em Journal of Machine Learning Research}, 21(140):1--67, 2020.

\bibitem{rajpurkar2016squad}
Pranav Rajpurkar, Jian Zhang, Konstantin Lopyrev, and Percy Liang.
\newblock Squad: 100,000+ questions for machine comprehension of text.
\newblock {\em arXiv preprint arXiv:1606.05250}, 2016.

\bibitem{robertson2009probabilistic}
Stephen Robertson and Hugo Zaragoza.
\newblock {\em The probabilistic relevance framework: BM25 and beyond}.
\newblock Now Publishers Inc, 2009.

\bibitem{seo2019real}
Minjoon Seo, Jinhyuk Lee, Tom Kwiatkowski, Ankur~P Parikh, Ali Farhadi, and
  Hannaneh Hajishirzi.
\newblock Real-time open-domain question answering with dense-sparse phrase
  index.
\newblock {\em arXiv preprint arXiv:1906.05807}, 2019.

\bibitem{shazeer2018adafactor}
Noam Shazeer and Mitchell Stern.
\newblock Adafactor: Adaptive learning rates with sublinear memory cost.
\newblock In {\em International Conference on Machine Learning}, pages
  4596--4604. PMLR, 2018.

\bibitem{singh2021end}
Devendra Singh, Siva Reddy, Will Hamilton, Chris Dyer, and Dani Yogatama.
\newblock End-to-end training of multi-document reader and retriever for
  open-domain question answering.
\newblock {\em Advances in Neural Information Processing Systems}, 34, 2021.

\bibitem{smith2022mtnlg}
Shaden Smith, Mostofa Patwary, Brandon Norick, Patrick LeGresley, Samyam
  Rajbhandari, Jared Casper, Zhun Liu, Shrimai Prabhumoye, George Zerveas,
  Vijay Korthikanti, et~al.
\newblock Using deepspeed and megatron to train megatron-turing nlg 530b, a
  large-scale generative language model.
\newblock {\em arXiv preprint arXiv:2201.11990}, 2022.

\bibitem{vaswani2017attention}
Ashish Vaswani, Noam Shazeer, Niki Parmar, Jakob Uszkoreit, Llion Jones,
  Aidan~N Gomez, {\L}ukasz Kaiser, and Illia Polosukhin.
\newblock Attention is all you need.
\newblock {\em Advances in neural information processing systems}, 30, 2017.

\bibitem{wu2022memorizing}
Yuhuai Wu, Markus~N Rabe, DeLesley Hutchins, and Christian Szegedy.
\newblock Memorizing transformers.
\newblock {\em arXiv preprint arXiv:2203.08913}, 2022.

\bibitem{xiong2020approximate}
Lee Xiong, Chenyan Xiong, Ye~Li, Kwok-Fung Tang, Jialin Liu, Paul Bennett,
  Junaid Ahmed, and Arnold Overwijk.
\newblock Approximate nearest neighbor negative contrastive learning for dense
  text retrieval.
\newblock In {\em International Conference on Learning Representations (ICLR)},
  April 2021.

\bibitem{xue-etal-2021-mt5}
Linting Xue, Noah Constant, Adam Roberts, Mihir Kale, Rami Al-Rfou, Aditya
  Siddhant, Aditya Barua, and Colin Raffel.
\newblock m{T}5: A massively multilingual pre-trained text-to-text transformer.
\newblock In {\em Proceedings of the 2021 Conference of the North American
  Chapter of the Association for Computational Linguistics: Human Language
  Technologies}, pages 483--498, Online, June 2021. Association for
  Computational Linguistics.

\bibitem{yang2019xlnet}
Zhilin Yang, Zihang Dai, Yiming Yang, Jaime Carbonell, Russ~R Salakhutdinov,
  and Quoc~V Le.
\newblock Xlnet: Generalized autoregressive pretraining for language
  understanding.
\newblock {\em Advances in neural information processing systems}, 32, 2019.

\bibitem{yogatama2021spalm}
Dani Yogatama, Cyprien de~Masson~d’Autume, and Lingpeng Kong.
\newblock Adaptive semiparametric language models.
\newblock {\em Transactions of the Association for Computational Linguistics},
  9:362--373, 2021.

\end{thebibliography}
\bibliographystyle{plain}

\clearpage

\section*{Checklist}

\begin{enumerate}

\item For all authors...
\begin{enumerate}
  \item Do the main claims made in the abstract and introduction accurately reflect the paper's contributions and scope?
    \answerYes{}
  \item Did you describe the limitations of your work?
    \answerYes{We mentioned the limitations in Section~\ref{sec:computation}.}
  \item Did you discuss any potential negative societal impacts of your work?
    \answerNo{}
  \item Have you read the ethics review guidelines and ensured that your paper conforms to them?
    \answerYes{}
\end{enumerate}

\item If you are including theoretical results...
\begin{enumerate}
  \item Did you state the full set of assumptions of all theoretical results?
    \answerNA{}
	\item Did you include complete proofs of all theoretical results?
    \answerNA{}{}
\end{enumerate}

\item If you ran experiments...
\begin{enumerate}
  \item Did you include the code, data, and instructions needed to reproduce the main experimental results (either in the supplemental material or as a URL)?
    \answerNo{We still need to clean up the code before it's ready.}
  \item Did you specify all the training details (e.g., data splits, hyperparameters, how they were chosen)?
    \answerYes{See Section~\ref{sec:lm-experiment} and Section~\ref{sec:nq}.}
	\item Did you report error bars (e.g., with respect to the random seed after running experiments multiple times)?
    \answerNo{It would take too long and too much resources to run each experiment multiple times.}
	\item Did you include the total amount of compute and the type of resources used (e.g., type of GPUs, internal cluster, or cloud provider)?
    \answerYes{We reported the type of device we used to train the model as well as the steps per second for mT5 XL model. For more information on run time, see Appendix.}
\end{enumerate}

\item If you are using existing assets (e.g., code, data, models) or curating/releasing new assets...
\begin{enumerate}
  \item If your work uses existing assets, did you cite the creators?
    \answerYes{}
  \item Did you mention the license of the assets?
    \answerYes{Will be mentioned in Appendix.}
  \item Did you include any new assets either in the supplemental material or as a URL?
    \answerNo{}
  \item Did you discuss whether and how consent was obtained from people whose data you're using/curating?
    \answerNA{}
  \item Did you discuss whether the data you are using/curating contains personally identifiable information or offensive content?
    \answerNA{}
\end{enumerate}

\item If you used crowdsourcing or conducted research with human subjects...
\begin{enumerate}
  \item Did you include the full text of instructions given to participants and screenshots, if applicable?
    \answerNA{}
  \item Did you describe any potential participant risks, with links to Institutional Review Board (IRB) approvals, if applicable?
    \answerNA{}
  \item Did you include the estimated hourly wage paid to participants and the total amount spent on participant compensation?
    \answerNA{}
\end{enumerate}

\end{enumerate}

\clearpage

\section*{Appendix}
\begin{appendices}
\section{Full data on POS tags and percentage of improvement on the log likelihood score.} \label{tab:pos_percent}

\begin{center}
\begin{tabular}{lrrrrr}
\toprule
{} &  \shortstack{Not in context,\\not in input} &  \shortstack{Not in context,\\in input} &  \shortstack{In context,\\not in input} &  \shortstack{In context,\\in input} &   Occurrence \\
\midrule
NN   &                      0.157771 &                  0.698062 &                 16.197878 &              5.578312 &  23639620 \\
NNP  &                      0.020258 &                  0.527689 &                 35.891013 &              8.141562 &  16840255 \\
IN   &                     -0.169446 &                 -0.187711 &                  3.308690 &              1.816983 &  13603903 \\
JJ   &                      0.400632 &                  1.011774 &                 13.077988 &              5.019096 &  12885333 \\
DT   &                      0.224373 &                  0.196791 &                  3.253798 &              2.453526 &  11128747 \\
NNS  &                      0.131760 &                  0.594879 &                 17.203110 &              5.242418 &  10088414 \\
.    &                      0.074150 &                 -0.292448 &                  7.026237 &              0.411417 &   6330229 \\
RB   &                      0.692010 &                  0.897788 &                  5.454908 &              2.711667 &   6174253 \\
VB   &                      0.384166 &                  0.659601 &                  9.219075 &              3.491015 &   5390427 \\
,    &                      0.731329 &                  0.710795 &                  3.420206 &              2.353697 &   5245713 \\
PRP  &                     -0.260621 &                 -0.006360 &                  3.305721 &              1.998650 &   5219071 \\
CC   &                      0.055603 &                  0.277403 &                  3.446472 &              2.634180 &   4463355 \\
VBZ  &                      0.355881 &                  0.093804 &                  5.872791 &              0.625349 &   3895160 \\
VBG  &                      0.588187 &                  0.913552 &                 11.268538 &              4.581329 &   3834853 \\
VBN  &                      0.260189 &                  0.650080 &                 12.332930 &              4.213668 &   3646850 \\
VBD  &                      0.312395 &                  0.378780 &                  6.878956 &              2.758350 &   3377739 \\
VBP  &                      0.333019 &                  0.362408 &                  5.038212 &              1.873432 &   3308882 \\
TO   &                      0.000127 &                 -0.183880 &                  3.921325 &              1.767642 &   3205519 \\
CD   &                      0.619997 &                  1.148265 &                 20.414486 &              5.950232 &   2723028 \\
PRP\$ &                      0.040851 &                  0.269473 &                  4.617825 &              2.683837 &   1942225 \\
MD   &                     -0.115006 &                  0.047148 &                  3.572827 &              2.559091 &   1590480 \\
:    &                     -0.729198 &                 -0.690925 &                  5.812855 &              3.404822 &    650535 \\
WDT  &                      0.455384 &                  0.571508 &                  3.444940 &              1.234093 &    576220 \\
WRB  &                      0.638024 &                  0.921696 &                  5.063341 &              3.163921 &    572072 \\
JJR  &                      1.010684 &                  1.612511 &                  7.880883 &              4.808698 &    483001 \\
(    &                      0.226634 &                  0.602507 &                  3.785168 &              2.695342 &    450846 \\
WP   &                      1.296407 &                  1.746595 &                  5.018384 &              3.681695 &    437709 \\
RP   &                      0.347023 &                  0.406551 &                  5.269570 &              3.544124 &    416556 \\
POS  &                     -1.028733 &                  0.472410 &                  8.667241 &              2.427072 &    387925 \\
)    &                      0.200041 &                  0.019679 &                  4.084305 &              1.586972 &    369449 \\
JJS  &                      0.499507 &                  0.790719 &                  7.975327 &              4.348411 &    342354 \\
NNPS &                     -0.206552 &                  0.327515 &                 30.412756 &              6.301600 &    233250 \\
EX   &                      0.793452 &                  1.222533 &                  4.670563 &              3.714079 &    201943 \\
RBR  &                      1.037655 &                  2.046677 &                  5.190104 &              4.050341 &    192250 \\
PDT  &                      0.604735 &                  1.509266 &                  3.752347 &              3.438381 &     84221 \\
RBS  &                      0.114155 &                  0.759014 &                  4.438339 &              4.140244 &     83746 \\
FW   &                      0.779163 &                  0.560080 &                 13.321369 &              4.776943 &     80379 \\
\$    &                     -0.071967 &                  0.671289 &                 18.502472 &              5.415212 &     63854 \\
UH   &                      0.154138 &                  0.207683 &                  7.501694 &              3.188253 &     45710 \\
''   &                      0.967397 &                  1.077652 &                 17.579383 &              2.770341 &     30181 \\
WP\$  &                      2.702986 &                  3.293384 &                  6.577276 &              3.215185 &     18004 \\
\#    &                      0.767495 &                  4.536762 &                 31.027920 &             14.052459 &     14969 \\
SYM  &                      0.343766 &                 -3.428490 &                 20.566550 &              3.729796 &      5532 \\
``   &                      1.352541 &                  4.927900 &                 35.315082 &             16.644833 &      2998 \\
LS   &                      3.417315 &                 -1.039910 &                 57.893078 &              8.573192 &       836 \\
\bottomrule
\end{tabular}
\end{center}

\section{Visualization of context and change in token log likelihood} \label{apdx:steps_per_sec}

\begin{figure}[H]
\begin{center}
\subfloat[Sampled from context-input-target triplet with top 1\% of context utility.]{
\includegraphics[width=0.9 \textwidth, trim={0 5cm 11cm 0}, clip]{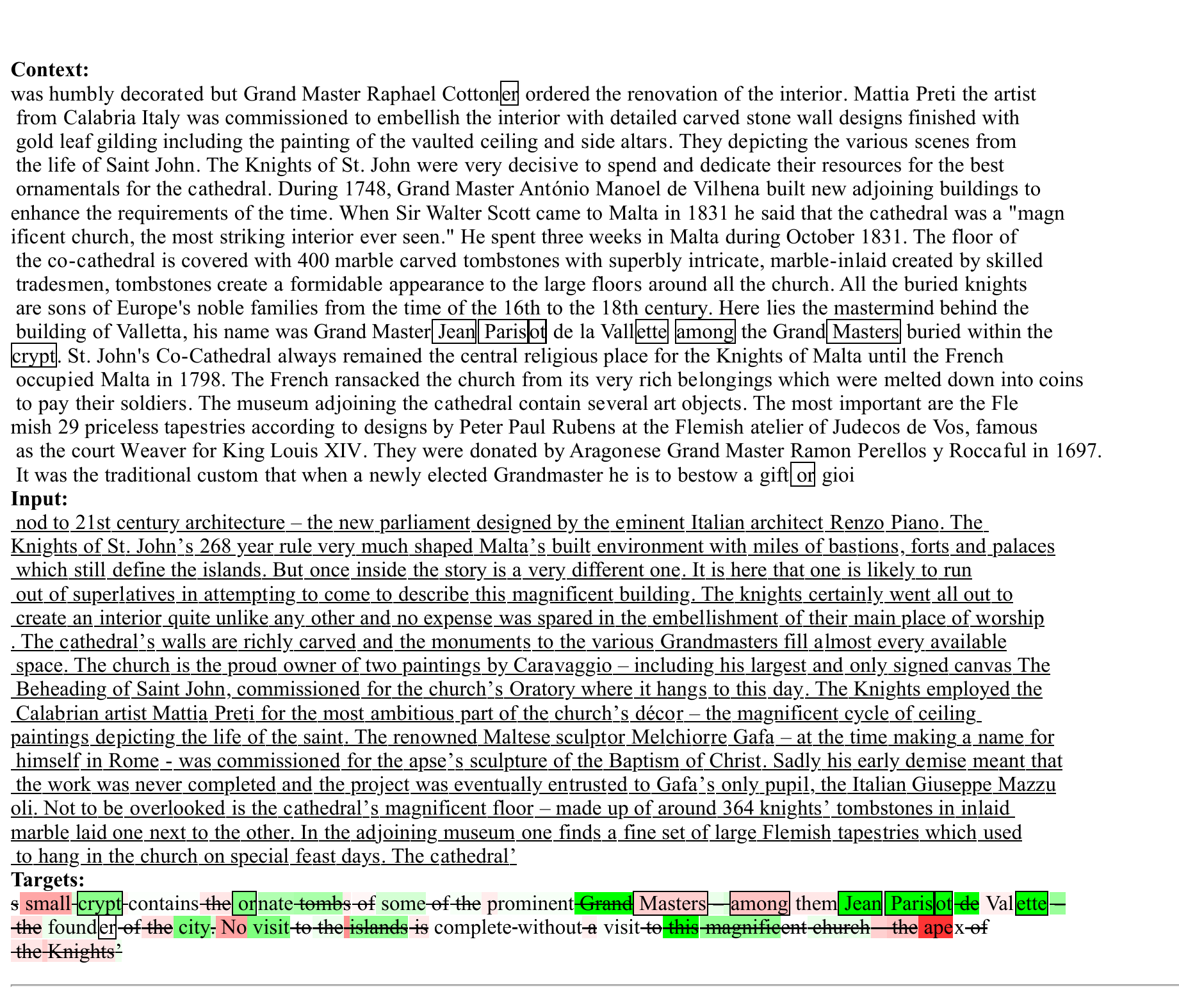}\label{fig:top1_pct}
}
\end{center}
\begin{center}
\subfloat[Randomly sampled.]{
\includegraphics[width=0.9 \textwidth, trim={0 8cm 11cm 0}, clip]{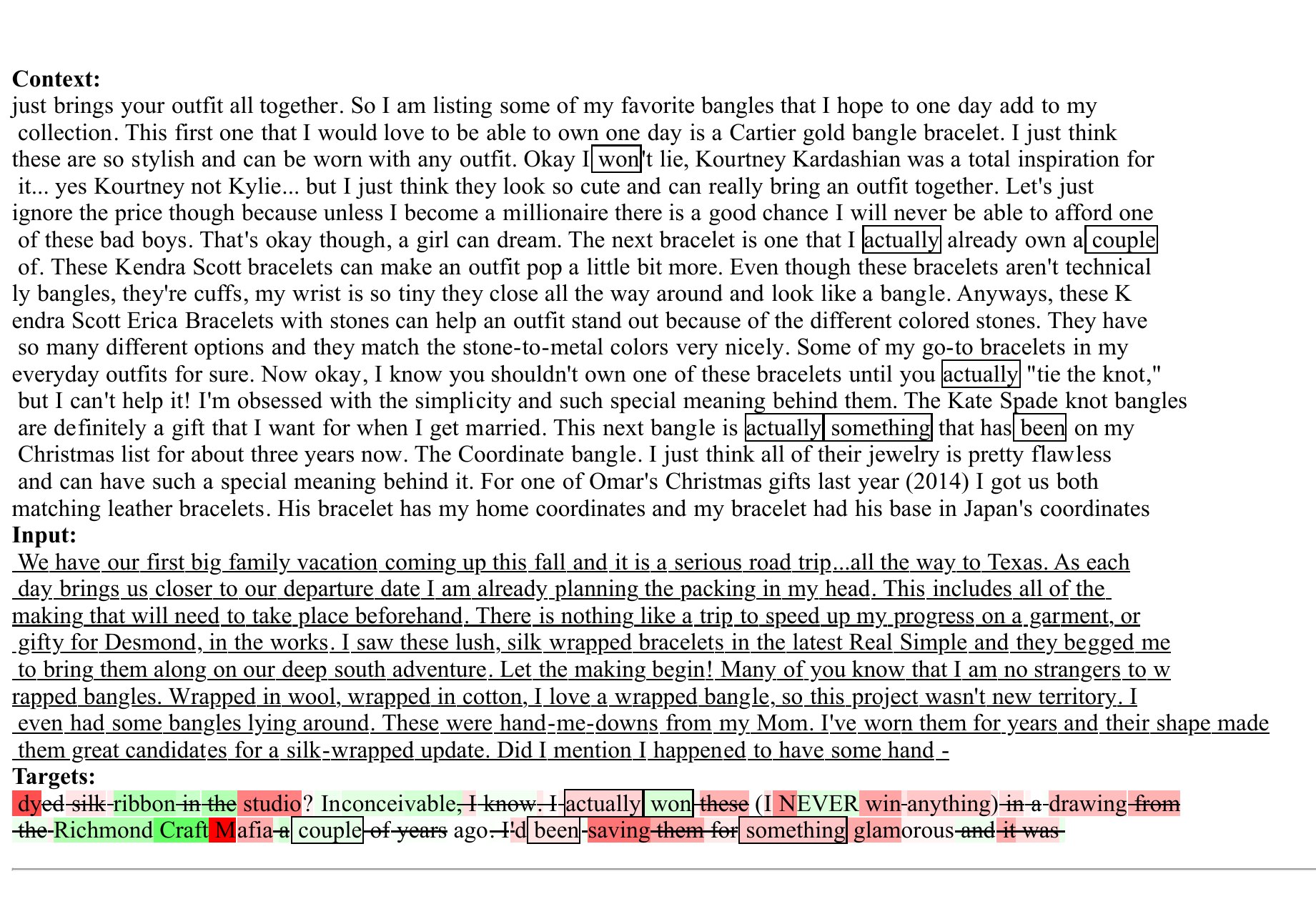}\label{fig:random}
}
\end{center}

\caption{Visualize the difference in log-likelihood for each target token with and without context.}

\end{figure}

Visualize the delta log-likelihood for each target token with and without context. The color of each target token is determined by \(\log P(y_i|y_{<i}, \x, \c, \mathbf{\theta}) - \log P(y_i|y_{<i}, \x, \mathbf{\theta'})\). The color is green if the difference is positive (context helped). The color is red if the difference is negative (context harmed). We put a box around a target token if it only appeared in the context, but not input. We put a strike-through on a token if it appeared in the input. We show two examples here. Figure~\ref{fig:top1_pct} is sampled from the context-input-target triplet with top 1\% of context utility. Figure~\ref{fig:random} is randomly sampled from the whole C4 train split. We provide more examples in the supplementary materials. They are interactive HTML files that you can mouse-over the tokens to see the difference in log-likelihood. 

\section{Model training steps per second} \label{apdx:steps_per_sec}

\begin{center}
\begin{tabular}{|c||c|c|c|c|}
\hline
Model Size & Steps/Sec & Training Devices & Number of Steps & Batch size \\
\hline
\hline
\multicolumn{5}{|c|}{C4 Language Modeling} \\
\hline
\texttt{113M} & 3.3 & 64 TPUv3 & 1,100,000 & 512 \\
\hline
\texttt{409M} & 2.2 & 64 TPUv3 & 1,100,000 & 512 \\
\hline
\texttt{1560M} & 1.5 & 128 TPUv3 & 1,100,000 & 512 \\
\hline
\hline
\multicolumn{5}{|c|}{Natural Question} \\
\hline
\texttt{409M} & 212.5 & 64 TPUv3 & 40,000 & 64 \\
\hline
\texttt{1560M} & 74.6 & 64 TPUv3 & 40,000 & 64 \\
\hline
\end{tabular}
\end{center}

\section{License for assets used in this work} \label{apdx:license}

\begin{center}
\begin{tabular}{|c||c|c|}
\hline
Name & License & Link \\
\hline
\hline
C4 English Dataset & ODC-By-1.0 & https://commoncrawl.org \\
\hline
Natural Question & CC BY-SA 3.0 & \shortstack{https://github.com/google-research-datasets/\\natural-questions/tree/master/nq\_open} \\
\hline
T5X & Apache-2.0 & https://github.com/google-research/t5x \\
\hline
mT5 & Apache-2.0 & https://github.com/google-research/multilingual-t5 \\
\hline
T5 Retrieval & Apache-2.0 & https://github.com/google-research/t5x\_retrieval \\
\hline
\end{tabular}
\end{center}

\end{appendices}

\end{document}